\documentclass[journal]{IEEEtran}
\usepackage[utf8]{inputenc} 
\usepackage[T1]{fontenc}    
\usepackage{hyperref}       
\usepackage{url}            
\usepackage{booktabs}       
\usepackage{amsfonts}       
\usepackage{nicefrac}       
\usepackage{microtype}      
\usepackage{graphicx}
\usepackage{subfigure}
\usepackage{booktabs}
\usepackage{dashrule}
\usepackage{color}
\usepackage{colortbl}
\definecolor{mygray}{gray}{.9}

\usepackage{multicol}

\usepackage{graphicx}
\usepackage{comment}
\usepackage{amsmath,amssymb} 
\usepackage{color}
\usepackage{mathrsfs}
\usepackage{url}
\usepackage{subfigure}
\usepackage{multirow}
\usepackage{bm}
\usepackage{amssymb}
\usepackage{animate}
\usepackage{multicol}
\usepackage{algorithmic}
\usepackage{algorithm}
\usepackage{amsmath}
\usepackage{caption}
\usepackage[numbers,sort&compress]{natbib}
\newcommand{\tabincell}[2]{\begin{tabular}{@{}#1@{}}#2\end{tabular}}
\usepackage{arydshln}

\newcolumntype{x}[1]{>{\centering\arraybackslash}p{#1pt}}

\newlength\savewidth

\newcommand{\ie}{\textit{i}.\textit{e}.}
\newcommand{\eg}{\textit{e}.\textit{g}.}
\ifCLASSINFOpdf
\else
\fi
\begin{document}
%
\title{Deep Motion Prior for Weakly-Supervised \\ Temporal Action Localization}
%
%
%

\author{Meng~Cao,
        Can Zhang,
        Long Chen,
        Mike Zheng Shou,
        Yuexian Zou\footnotemark$^\dagger$
\thanks{Meng Cao, Can Zhang, and Yuexian Zou are with School of Electronic and Computer Engineering, Peking University, Shenzhen, China. Yuexian Zou is also with the Peng Cheng Laboratory, Shenzhen, China. Long Chen is with the Department of Electrical Engineering, Columbia University, New York, USA. Mike Zheng Shou is with National University of Singapore, SG. $^\dagger$ Yuexian Zou is the corresponding author. }
}

%
%

\markboth{Journal of \LaTeX\ Class Files,~Vol.~14, No.~8, August~2015}%
{Shell \MakeLowercase{\textit{et al.}}: Bare Demo of IEEEtran.cls for IEEE Journals}
%



\maketitle
\renewcommand{\thefootnote}{\fnsymbol{footnote}}

\begin{abstract}\label{abs}
Weakly-Supervised Temporal Action Localization (WSTAL) aims to localize actions in untrimmed videos with only video-level labels. Currently, most state-of-the-art WSTAL methods follow a Multi-Instance Learning (MIL) pipeline: producing snippet-level predictions first and then aggregating to the video-level prediction. However, we argue that existing methods have overlooked two important drawbacks: 1) inadequate use of motion information and 2) the incompatibility of prevailing cross-entropy training loss. In this paper, we analyze that the motion cues behind the optical flow features are complementary informative. Inspired by this, we propose to build a context-dependent motion prior, termed as \textit{motionness}. Specifically, a motion graph is introduced to model motionness based on the local motion carrier (\textit{e.g.,} optical flow). In addition, to highlight more informative video snippets, a motion-guided loss is proposed to modulate the network training conditioned on motionness scores. Extensive ablation studies confirm that motionness efficaciously models action-of-interest, and the motion-guided loss leads to more accurate results. Besides, our motion-guided loss is a plug-and-play loss function and is applicable with existing WSTAL methods. Without loss of generality, based on the standard MIL pipeline, our method achieves new state-of-the-art performance on three challenging benchmarks, including THUMOS'14, ActivityNet v1.2 and v1.3. 
\end{abstract}


\section{Introduction}\label{sec:intro}
\begin{figure}[!t]
	\begin{center}
		\includegraphics[width=\linewidth]{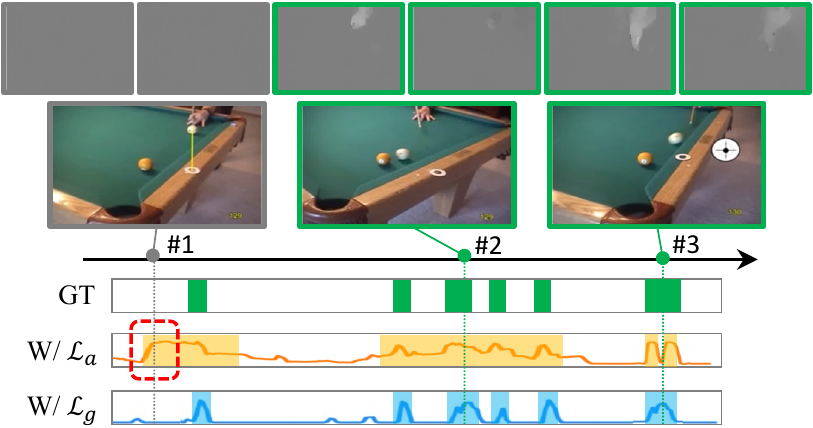}
	\end{center}
	\caption{Comparisons between TCAS distributions (for category \texttt{Billiards}) of the MIL baseline with XE loss (W/$\mathcal{L}_{a}$) and proposed motion-guided loss (W/$\mathcal{L}_{g}$). Prediction results are obtained by thresholding and visualized by the corresponding mask. GT is short for ground truth action instances of category \texttt{Billiards}. We select three representative frames and show their optical flow maps in horizontal and vertical directions.}
	\label{fig:teaser}
\end{figure}

\IEEEPARstart{T}{emporal} Action Localization (TAL) aims at identifying the start and end timestamps of all the action instances occurring in an untrimmed video. It is an indispensable building block for numerous video understanding applications such as intelligent video summarization~\cite{lee2012discovering}, surveillance analysis~\cite{vishwakarma2013survey}, video retrieval~\cite{hu2011survey}, and video grounding~\cite{chen2020rethinking,lu2019debug,xiao2021boundary,cao2021pursuit,xiao2021natural,cao2022correspondence} \emph{etc}. However, the required frame-level labeling is subjective, labor-intensive, and error-prone. Therefore, Weakly-Supervised Temporal Action Localization (WSTAL) has gained intensive attention, where the ``weak" denotes a much cheaper labeling cost at the video level. For example, as shown in Fig.~\ref{fig:teaser}, given the untrimmed video and category \texttt{Billiards}, WSTAL models aim to detect the temporal locations of all \texttt{Billiards} action instances. Unfortunately, there is no free lunch --- it is essentially challenging to perform frame-wise localization and classification with only the video-level supervision, especially for complex visual scenes.

The \textit{de} \textit{facto} paradigm of most state-of-the-art WSTAL methods~\cite{wang2017untrimmednets,nguyen2018weakly,paul2018w,narayan20193c,lee2020background,zhang2021synergic} is to first pre-extract video RGB and optical flow features from pre-trained vision models (\eg, C3D~\cite{tran2015learning} and I3D~\cite{carreira2017quo}), then follow the Multi-Instance Learning (MIL) principle. Specifically, snippet-wise classifications are performed over time to generate Temporal Class Activation Sequence (TCAS), which is then aggregated to predict the video-level classification score. The whole process is optimized by a standard cross-entropy (XE) loss. We call this framework \emph{baseline} in our paper.

Although great progress has been made in this direction, two main drawbacks still exist in today's WSTAL methods: \textit{(i) Inadequate use of motion information:} We argue that the motion cue behind the optical flow modality is empirically more informative (cf. Sec.~\ref{subsec:ana_dmp}) and should not be treated equally as the RGB modality. Most existing methods, however, fail to extensively explore the motion information behind optical flow. Specifically, current WSTAL methods always resort to either \emph{early fusion} or \emph{late fusion} manner. For early fusion methods~\cite{paul2018w,shou2018autoloc,liu2019weakly}, the extracted RGB and optical flow features are concatenated before feeding to the network. For late fusion methods~\cite{nguyen2018weakly,zhai2020two}, they conduct predictions on each modality independently and then fuse the results by the weighted sum or other post-processing steps. However, this simple concatenation or fusion manners are indirect and inadequate, leading to many false detection results. For example, as shown in Fig.~\ref{fig:teaser}, frame\footnote{We slightly abuse "frame" here, and we refer to it as a video snippet.\label{frame_path}} \#1 shares similar appearances with frame \#2, but it is actually a background frame since it is only a stationary frame with explanatory narrations. In this case, the motion information can help us to easily distinguish them because their optical flow features have low responses, which implies that this frame is unlikely to be an action. \textit{(ii) Incompatibility of XE loss:} XE loss is designed to measure the performance of a classification model and is inherently incompatible with our localization task. Specifically, the XE loss encourages the discriminative video clips that tend to be fragmentary without covering the entire actions~\cite{singh2017hide,yuan2019marginalized,zhong2018step,liu2019completeness}. For example, frame \#3 is within a ground truth interval, yet it is misclassified and leads to incomplete localization results. Similarly, let's focus on the motion information. From the optical flow images, we find similar patterns between frame \#3 and frame \#2. Thus, introducing coherent motion information in the loss function will lead to more complete and accurate predictions.

Motivated by this, we propose \textbf{D}eep \textbf{M}otion \textbf{P}rior Network (DMP-Net) to make full use of the optical flow modality via learning an effective context-dependent motion representation (referred to as \emph{motionness} in our paper). Our motionness is with global perception and focuses on action-of-interest regardless of the background and irrelevant motions. Based on this, we propose a motion-guided loss, which is a plug-and-play loss function that may be an alternative to the traditional XE loss under the weakly-supervised setting.

For the motionness modeling, we introduce a motion graph to enlarge the receptive field of each temporal snippet since optical flow is the local motion representation calculated between two consecutive frames. Specifically, this paper investigates the temporal relationships from two perspectives: \emph{positional relationship} and \emph{semantic relationship}. To illustrate this, let's revisit frame \#3 in Fig.~\ref{fig:teaser}. \textbf{1)} positional relationship: frame \#3 is misclassified while its surrounding frames are all correctly predicted. These adjacent snippets will provide contextual information, which is advantageous for correct classification. \textbf{2)} semantic relationship: frame \#2 is distant from frame \#3, but they share the similar motion patterns (similar \texttt{billiard} ball hitting processes). Thus, it provides indicative hints for frame \#3 and leads to more comprehensive information. Based upon these two relationships, we construct the motion graph which encourages both the adjacent positional contexts and the distant semantic correlations. 

For motion-guided loss, we aim to use the modeled motionness to modulate the network training. We also start from an intuitive idea that the higher motionness of one timestamp, the greater the probability of it becoming action-of-interest. Let's recall the video-level classification aggregation process in the baseline. Following the top-\emph{k} mean strategy in~\cite{lee2020background,nguyen2019weakly,zhang2021cola,paul2018w,cao2022locvtp,zhang2022unsupervised}, for each category, the mean value of \emph{k} terms with the largest TCAS values is computed as the video-level classification score. In this paper, we further evaluate the motion characteristics of these selected terms. Specifically, we take the values of the corresponding terms in the motionness sequence and incorporate these values into the loss computation. In this way, terms with both high TCAS and motionness scores are highlighted while terms with low motionness scores are down-weighted. Experimental results have shown that this intuitive design leads to better results.

In summary, we make three contributions in this paper:
\begin{itemize}
  \item We argue that a context-dependent deep motion prior is crucial in accurate action localization and we obtain it by applying a motion graph to exploit the relationships between temporal nodes.
  \item An efficient motion-guided loss is developed to inform the whole pipeline of more motion cues, which can be seamlessly plugged into any existing WSTAL models.
  \item Extensive experiments on three challenging datasets have demonstrated the effectiveness of our proposed DMP-Net.
\end{itemize}

\section{Related Work}\label{relatedwork}
	\begin{figure*}[t]
		\begin{center}
			\includegraphics[width=0.95\linewidth]{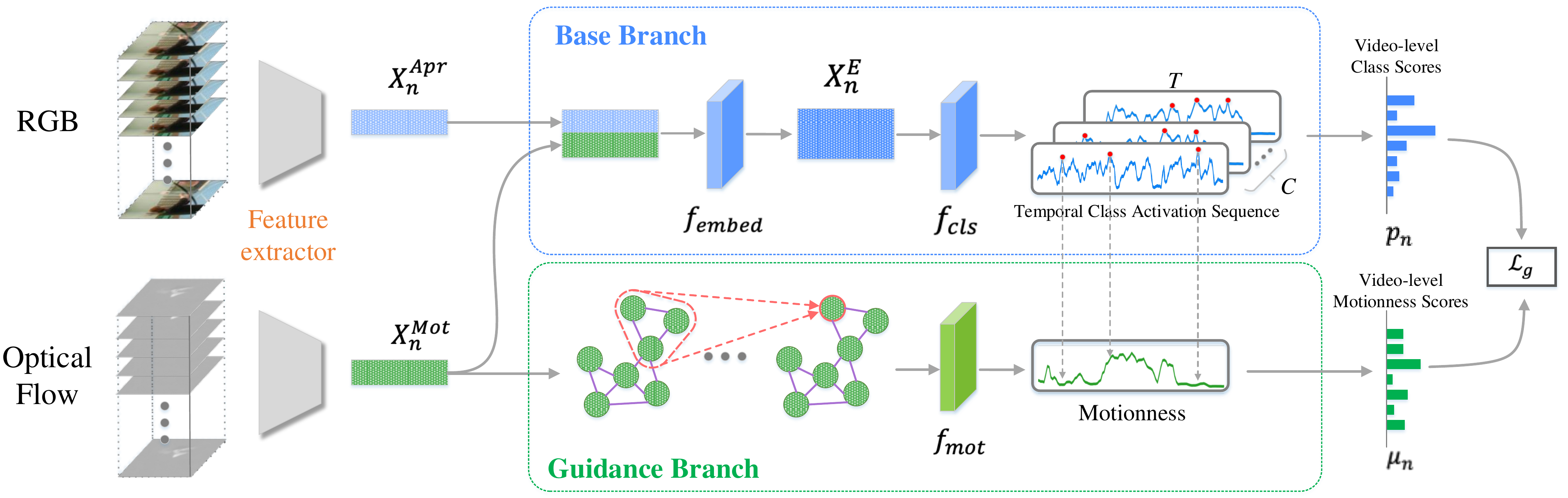}
		\end{center} 
		\caption{Schematic illustration of the proposed DMP-Net, which consists of two branches: (a) Base branch to produce class-specific probabilities (TCAS) and (b) Guidance branch to output class-agnostic deep motion prior. In the base branch, for each channel (category) of TCAS, top-\emph{k} terms with largest values (marked as red nodes) are selected to aggregate the video-level classification results. In the guidance branch, the corresponding item in the motionness sequence is also selected and fed to our Motion-guided Loss $\mathcal{L}_{g}$. For clarity, we only show the motionness selection for the first channel and the remaining channels are similar.}
		\label{fig:short}
	\end{figure*}

\noindent\textbf{Weakly-Supervised Temporal Action Localization.} WSTAL requires only video-level labels and saves large human effort from the frame-level labeling. The primer work UntrimmedNets~\cite{wang2017untrimmednets} formulates this problem as a Multi-Instance Learning (MIL) framework by evaluating the contribution of each clip to the video classification. Later, STPN~\cite{nguyen2018weakly} applies this attention mechanism to the feature level with a proposed sparsity constraint. To regularize the feature representation, W-TALC~\cite{paul2018w} introduces deep metric learning as a complement. Inherently, attention-MIL methods tend to produce incomplete localization results. To tackle this, several works~\cite{singh2017hide,liu2019completeness,zeng2019breaking} try to extend the discriminative regions via randomly hiding patches or suppressing the dominant response. To model complete actions, Liu \textit{et al.}~\cite{liu2019completeness} develop a parallel multi-branch classification architecture with the help of the generated hard negative data. Zeng \textit{et al.}~\cite{zeng2019breaking} proposes an iterative training strategy, which selects the most discriminative action instances in each iteration and removes them in the next iteration. For the clear distinction between background and foreground, several background modeling works~\cite{nguyen2019weakly,lee2020background,shi2020weakly} are proposed to ease the action-context confusion. CleanNet~\cite{liu2019weakly} introduces additional pseudo-supervision by leveraging the temporal contrast in snippet-level action classification predictions. UM-Net~\cite{lee2021weakly} models the background frames as out-of-distribution samples and realizes this uncertainty learning via multiple instance learning. SF-Net~\cite{ma2020sf} introduces extra supervisions by annotating one single frame within the action and mines pseudo action and background frames based on this annotation.

Typically, most WSTAL methods are based on the extracted RGB and optical flow with two possible fusion ways. Early fusion methods~\cite{paul2018w,shou2018autoloc,liu2019weakly,lee2020background} concatenate two modalities before feeding to the network while late fusion methods~\cite{nguyen2018weakly,nguyen2019weakly} compute a weighted sum of their respective outputs. Either of the two fusion strategies treats RGB and optical flow equally or independently. However, the motion cue behind the optical flow modality is empirically more informative. Experimental results in previous publications~\cite{zhai2020two} and our experiments (Sec.~\ref{subsec:ana_dmp}) have also demonstrated that when using the single modality, optical flow based methods are superior to RGB based ones. Therefore, we reuse optical flow to model a context-dependent motion prior and use it to guide the network training.

\noindent\textbf{Graph Convolutional Networks.} GCN is firstly proposed in \cite{kipf2016semi} for non-Euclidean structures. Recently, GCNs have been successfully applied to multiple computer vision tasks including video understanding~\cite{yan2018spatial}, object detection~\cite{xu2019spatial}, and point cloud segmentation~\cite{wang2019graph}. PGCN~\cite{zeng2019graph} uses GCNs to explore the relations between proposals to refine the boundary regression. G-TAD~\cite{xu2020g} incorporates multi-level video context into feature representations and casts action detection as a sub-graph localization problem. For visual concept localization~\cite{jiang2019class}, Zhu \emph{et al.}~\cite{zhu2021few} propose to conduct dynamic feature relation reasoning via a graph convolutional network on the extracted region features. Similarly, we apply GCN on generated motionness sequence to enhance the interactions between each snippet.

Action Graph~\cite{rashid2020action} also builds a graph to model snippet-level relationships. Our method, however, differs from \cite{rashid2020action} in the following two aspects. Firstly, in \cite{rashid2020action}, the graph convolution is applied on the concatenation of the RGB and optical flow features while our DMP-Net focuses on the effective motion modeling. Secondly, \cite{rashid2020action} designs a dense graph, \ie, every node pair is connected. In our experiments, we find that this fully connected way suffers from the feature slowness~\cite{zhang2012slow} in videos, making the learned weight focus on surrounding snippets. In contrast, our sparse graph with positional edges and semantic edges effectively captures both the contextual and long-term semantic information.

\section{Method}\label{methods}

In this section, we first present the problem formulation in Sec.~\ref{sec:feature}. We present the general scheme of our DMP-Net in Sec.~\ref{sec:overall}. Then we detail the modeling of the base branch and the guidance branch in Sec.~\ref{sec:action} and Sec.~\ref{subsec: motionness}, respectively. Based on them, we introduce a motion-guided loss in Sec.~\ref{sec:loss}. Finally, the inference process is given in Sec.~\ref{sec:infer}.

\subsection{Notations and Preliminaries} \label{sec:feature}
In the training phase, assume that we are given a set of $N$ untrimmed videos $\{V_n\}_{n=1}^{N}$ and their video-level labels $\{\bm{y}_n\}_{n=1}^{N}$, where $\bm{y}_n \in \mathbb{R}^C$ is a multi-hot vector, and $C$ is the number of action categories. Following the common practice \cite{nguyen2018weakly,nguyen2019weakly,lee2020background}, we represent each video with fixed-length non-overlapping snippets, \textit{i.e.,} $V_n = \{ S_{n,t} \}_{t=1}^{T}$, where $T$ denotes the number of sampled snippets. Then the appearance features $\bm{X}^{Apr}_n=\{ \bm{a}_t \}_{t=1}^{T}$ and motion features $\bm{X}^{Mot}_n=\{ \bm{m}_t \}_{t=1}^{T}$ are extracted with the pre-trained feature extractor (\emph{e.g.}, I3D \cite{carreira2017quo}) from the sampled RGB snippets and optical flow snippets, respectively. Here, $\bm{a}_t \in \mathbb{R}^{d}$, $\bm{m}_t \in \mathbb{R}^{d}$ and $d$ is the feature dimension of each snippet.

\subsection{Overview of DMP-Net} \label{sec:overall}
We try to emphasize the importance of the optical flow modality over the RGB one and solve the incompatibility problem of cross-entropy loss. Thus, we propose a general motion-guided loss that may replace the traditional cross-entropy loss in the existing WSTAL methods in a plug-and-play manner. To demonstrate the effectiveness of the proposed loss, we design an effective WSTAL model called DMP-Net, which consists of two branches: the base branch and the guidance branch. The whole network is optimized with our motion-guided loss. The overall pipeline of our DMP-Net is demonstrated in Fig.~\ref{fig:short}. 

Without loss of generality, we take the standard multi-instance learning pipeline as the base branch. Specifically, the RGB and optical flow features are concatenated to generate snippet-level classification results, \ie, Temporal Class Activation Sequence (TCAS)~\cite{paul2018w}. For the guidance branch, we seek to build a context-dependent motion representation (termed as motionness) based on the optical flow features. To this end, a graph convolutional module is introduced to model snippet-level relations. Instead of directly constructing a dense fully-connected graph, we carefully design a computation-efficient graph with sparse edge connections. Here we take two types of relations, \ie, \emph{positional edges} to utilize neighborhood correlations and \emph{semantic edges} to capture semantically related but disjointed snippets. Finally, the generated motionness is used to guide the network training in our motion-guided loss.

\subsection{Base Branch} \label{sec:action}

The base branch is shown in the top part of Fig.~\ref{fig:short}. Firstly, we apply an embedding function $f_{embed}$ over the concatenation of $\bm{X}^{Apr}_n$ and $\bm{X}^{Mot}_n$ to obtain the features $\bm{X}_n^E \in \mathbb{R}^{T \times 2d}$. $f_{embed}$ is implemented with a temporal convolution followed by the ReLU activation function. Given the embedded features $X_n^E$, we apply a classifier $f_{cls}$ to obtain snippet-level class scores, namely Temporal Class Activation Sequence.
\begin{equation}
	\mathcal{A}_n = f_{cls}(X_n^E; \phi_{cls}),
	\label{equation:action_cls}
\end{equation}
\noindent where $f_{cls}$ contains a temporal convolution followed by ReLU activation while $\phi_{cls}$ are the learnable parameters. The obtained $\mathcal{A}_n \in \mathbb{R}^{T \times C}$ represents the action classification results occurring at each temporal snippet.

\subsection{Guidance Branch} \label{subsec: motionness}

In this branch, we aim to model a context-dependent motion prior called motionness. Traditionally, the optical flow feature is widely adopted to provide temporal motion information~\cite{simonyan2014two,rashid2020action,zhai2020two}. However, optical flow has an inherent disadvantage that it can only reflect the local motion information since it is computed between two consecutive frames. To obtain an effective motion prior, we build a motion graph to model the snippet-snippet relations and thus eliminate interfering motion information (\eg, background or unrelated motions). Next, we present our motion graph construction process and motionness modeling with GCNs sequentially.

\textbf{Motion Graph Construction.} 
For the graph construction, one possible way is to build a dense fully connected graph, \ie, each snippet is connected with the other snippets across the whole video. This intuitive way suffers from the two drawbacks: \emph{(i)} The dense connection way is expensive and the number of edges is the quadratic order of the number of snippets; \emph{(ii)} Since the short-term video varies slowly, this \emph{slowness} prior~\cite{zhang2012slow} leads to the large feature similarity between adjacent snippets. Thus the feature updating within the dense graph is dominated by surrounding snippets because of the similar feature representations and the faraway snippets are ignored.

To alleviate this, we build a motion graph with sparse edge connections and encourage both the positionally adjacent snippets and the semantically related but disjointed snippets. Formally, let $\mathcal{G}=\{\mathcal{M}, \mathcal{E}\}$ be the graph of $T$ nodes with the node set $\mathcal{M}$ and edge set $\mathcal{E}$. Furthermore, the adjacency matrix associated with $\mathcal{G}$ is denoted as $\bm{G} \in \mathbb{R}^{T \times T}$. For the graph $\mathcal{G}$, each node (\textit{i.e.}, $\bm{m}_i \in \mathbb{R}^{d}$) is instantiated as the optical flow features of the corresponding snippet. Here, we devise two types of edge construction approaches by exploiting both the positional awareness and the semantic similarities, \ie, $\mathcal{E}=\mathcal{E}_{pos} \cup \mathcal{E}_{smt}$.

\begin{figure}[t]
	\begin{center}
		\includegraphics[width=\linewidth]{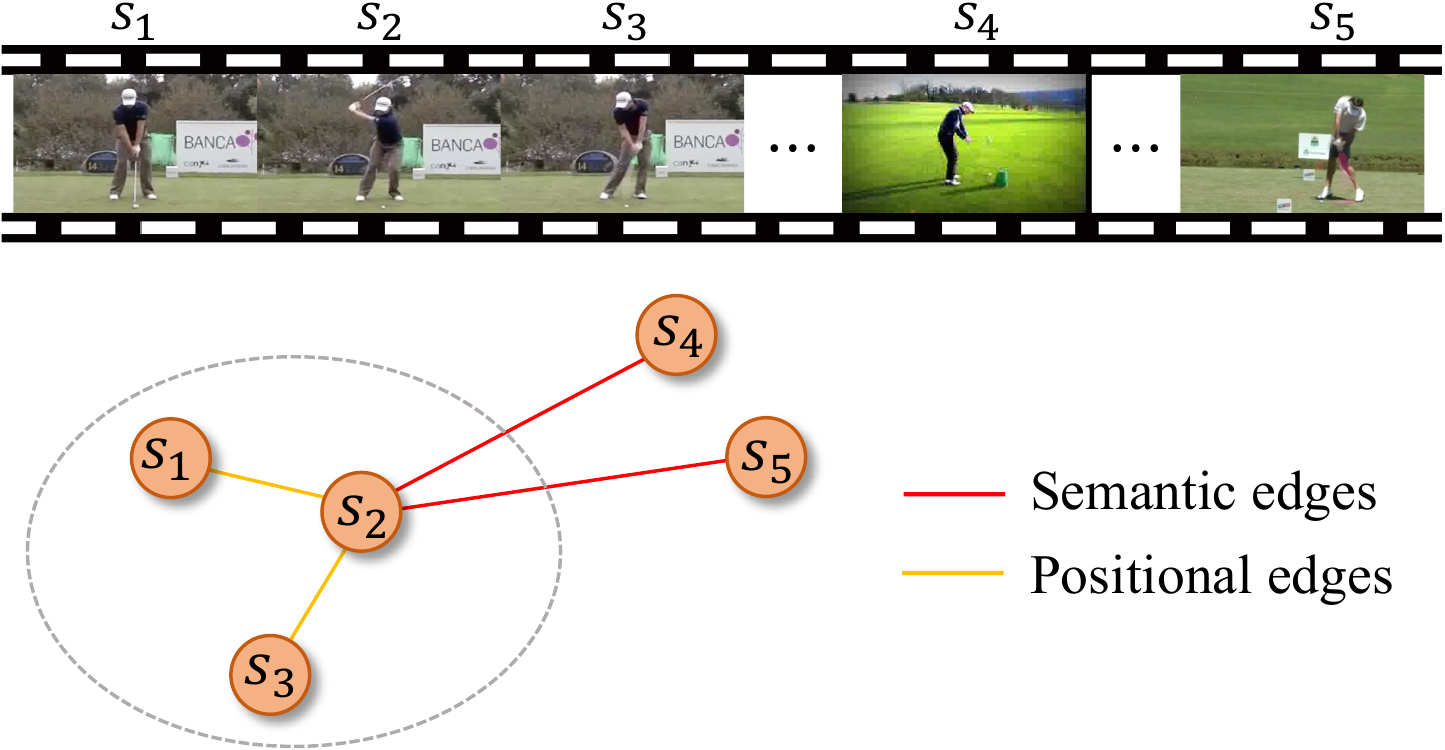}
	\end{center}
	\caption{Schematic illustration of the motion graph. We demonstrate the edge connections from the temporal node $s_2$. Note that we show the RGB images while the actual graph convolution is conducted on the optical flow modality.}
	\label{fig:graphBuild}
\end{figure}

\textbf{\textit{Positional edges $\mathcal{E}_{pos}$.}} Snippets that are close in location have a natural temporal connection and the message passing among them will facilitate the feature representation. As the example shown in Fig.~\ref{fig:graphBuild}, $s_1$, $s_2$ and $s_3$ are the different stages of the \texttt{GolfSwing} action. Thus $s_1$ and $s_3$ provide fruitful contextual information for the modeling of $s_2$. Generally, we establish an edge between snippet $\bm{m}_i$ and $\bm{m}_j$ if their temporal distance is below one pre-set threshold:

\begin{equation}
\mathcal{E}_{pos}=\left\{e_{ij} \bigm| \frac{\left|\bm{t}_i - \bm{t}_j\right|}{T} < \theta_{pos}\right\}
\end{equation} 
\noindent where $\bm{e}_{ij} := \langle\bm{m}_i, \bm{m}_j\rangle$ is the edge connecting $\bm{m}_i$ and $\bm{m}_j$; $\bm{t}_i$ and $\bm{t}_j$ are the temporal indices of snippet $\bm{m}_i$ and $\bm{m}_j$, respectively. $\theta_{pos}$ is a certain threshold. In this way, each snippet is enhanced with its contextual information, which is obviously helpful to refine the feature representation and increase the localization accuracy.

\textbf{\textit{Semantic edges $\mathcal{E}_{smt}$.}} Besides the temporal context information, the semantic correlation is also beneficial for each snippet even when they are scattered in time. Note that the untrimmed video often contains multiple action segments. Thus, finding the action instances belonging to the same or similar categories will enhance the discriminative motion patterns. As shown in Fig.~\ref{fig:graphBuild}, $s_4$ and $s_5$ are both the golf swing actions across the different scenes and actors. Despite the long temporal distance, $s_4$ and $s_5$ share similar semantic information with $s_2$, and this kind of connection is beneficial for the motionness modeling. Thus we set the semantic edges for long-range correlation modeling, which helps build high-level and global-aware relations for the same or similar action instances within one video. In particular, we use the cosine similarity between snippet nodes to find semantically related nodes, \ie,
\begin{equation}
\small
\mathcal{E}_{smt}=\left\{e_{ij} \mid \frac{\left|\bm{t}_i - \bm{t}_j\right|}{T} > \theta_{pos}; 
\frac{(\mathbf{W}_{1}\bm{m}_{i})^{\top} (\mathbf{W}_{2}\bm{m}_{j})}{\left\|\mathbf{W}_{1}\bm{m}_{i}\right\|_{2} \cdot\left\|\mathbf{W}_{2}\bm{m}_{j}\right\|_{2}} > \gamma\right\},
\end{equation}
\noindent where $\gamma$ is the similarity threshold. $\mathbf{W}_{1}, \mathbf{W}_{2} \in \mathbb{R}^{d \times d}$ are learnable parameters.

\textbf{Motionness Modeling.}
Given the constructed graph, we apply $K$-layer GCNs on the constructed motion graph to perform reasoning. GCNs facilitate the message passing of the graph and update motion features for each snippet node. Concretely, for the $k$-th layer:
\begin{equation}
	\bm{X}^{k}=\bm{G}\bm{X}^{k-1}\bm{W}^k, 1 \leq k \leq K, \label{equ:gcn_motion}
\end{equation}
\noindent where $\bm{W}^k \in \mathbb{R}^{d_k \times d_k}$ is the learnable parameter matrix; $\bm{X}^k \in \mathbb{R}^{T \times d_k}$ are the hidden motion features for all snippets at layer $k$; $\bm{X}^0 = \{ \bm{m}_i \}_{i=1}^{T} \in \mathbb{R}^{T \times d}$ are the input motion features. $\bm{G}$ is the adjacent matrix formulated as follows:
\begin{equation}
\bm{G}_{i j}= \begin{cases}\frac{\bm{m}_{i}^{\top} \bm{m}_{j}}{\left\|\bm{m}_{i}\right\|_{2} \cdot\left\|\bm{m}_{j}\right\|_{2}}, & e_{i j} \in \mathcal{E}; \\ 0, & e_{i j} \notin \mathcal{E}.\end{cases}
\label{equ:adj_def}
\end{equation}

Note that following the common practice~\cite{zeng2019graph,xu2020g}, a short-cut path is used to preserve the input features and the final output is as follows:
\begin{equation}
	\bm{X}^{K}=\bm{X}^{K} \| \bm{X}^{0},
\end{equation}
where $\|$ denotes the concatenation operation. The graph convolutions allow the network to compute the response of a node based on its edges defined by the graph, thereby enlarging the receptive field and facilitating information exchange among neighboring or distant snippets. 

After obtaining the updated motion features $\bm{X}^{K}$, we apply a binary classifier $f_{mot}$ to obtain the motionness. Specifically, the classifier contains a temporal convolution followed by the ReLU activation function. This can be formulated as follows:
\begin{equation}
	\mathcal{M}_n=f_{mot}(\bm{X}^{K};\phi_{mot}),
\end{equation}
\noindent where $\phi_{mot}$ is the learnable parameter. The obtained $\mathcal{M}_n \in \mathbb{R}^{T}$ represents the motionness scores for each temporal snippet.

\subsection{Motion-guided Loss}  \label{sec:loss}

Before we specify our motion-guided loss, let’s revisit the commonly-adopted video-level classification loss, which is in a traditional binary cross-entropy form.
	
To get the video-level class scores, we aggregate snippet-level class scores computed in Eq.~\eqref{equation:action_cls}. Following~\cite{wang2017untrimmednets,paul2018w,lee2020background}, we take the top-\emph{k} mean strategy: for each class $c$, we take $\lfloor \frac{T}{r} \rfloor$ ($r$ is the selection ratio) terms with the largest class-specific TCAS values and compute their means as $a_{n;c}$, namely the video-level class score for class $c$ of video $V_n$. The index set of the corresponding selected elements is denoted as $\mathcal{S}_{n;c}^{a}$. After obtaining $a_{n;c}$ for all the $C$ classes, we apply Softmax function on $a_n$ along the class dimension to get the video-level class possibilities $p_n \in \mathbb{R}^{C}$, namely \(p_n=\text{softmax}(a_n)\). XE loss ($\mathcal{L}_{a}$) is then calculated in the cross-entropy form: 
\begin{equation}
	\mathcal{L}_{a} = -\frac{1}{N} \sum^{N}_{n=1} \sum^{C}_{c=1} \hat{y}_{n;c}  \mathrm{log} (p_{n;c}),
	\label{equation:action_loss}
\end{equation}
where $\hat{y}_n \in \mathbb{R}^{C}$ is the normalized ground-truth.

XE loss is only associated with the TCAS sequence, which is modeled based on the motion and appearance feature concatenation. In previous discussions, we have noticed that motion information provides more hints to facilitate the localization. Thus the commonly-adopted top-\emph{k} mean strategy is a rather coarse aggregation manner since it does not consider the motionness score for each selected term. To alleviate this, we additionally generate the video-level motionness scores $\mu_{n}=\{\mu_{n;c}\} \in \mathbb{R}^{C}$ as the mean motionness value for those snippets with top-\emph{k} actionness values:
\begin{equation}
\mu_{n;c} = \text{mean}\{ \mathcal{M}_{n;t} | t \in \mathcal{S}_{n;c}^{a}\}.
\label{equation:mn}
\end{equation}
where $\mathcal{S}_{n;c}^{a}$ is the index set of the top-\emph{k} terms for category $c$ in the TCAS sequence as previously defined.

Intuitively, the network should highlight snippets that are simultaneously correctly classified and have a high motionness score, \ie, the motion-guided loss is: 
\begin{equation}
    \mathcal{L}_{g}= -\frac{1}{N} \sum^{N}_{n=1}  \sum^{C}_{c=1} {\mu_{n;c}}^{2} \hat{y}_{n;c}  \mathrm{log} (p_{n;c}) - \log {{\mu_{n;c}}^{2}},     
    \label{equation:guide_loss}
\end{equation}
where $\hat{y}_n$ is the same as previously defined in Eq.~\eqref{equation:action_loss}.

\begin{figure}[t]
	\begin{center}
		\includegraphics[width=0.9\linewidth]{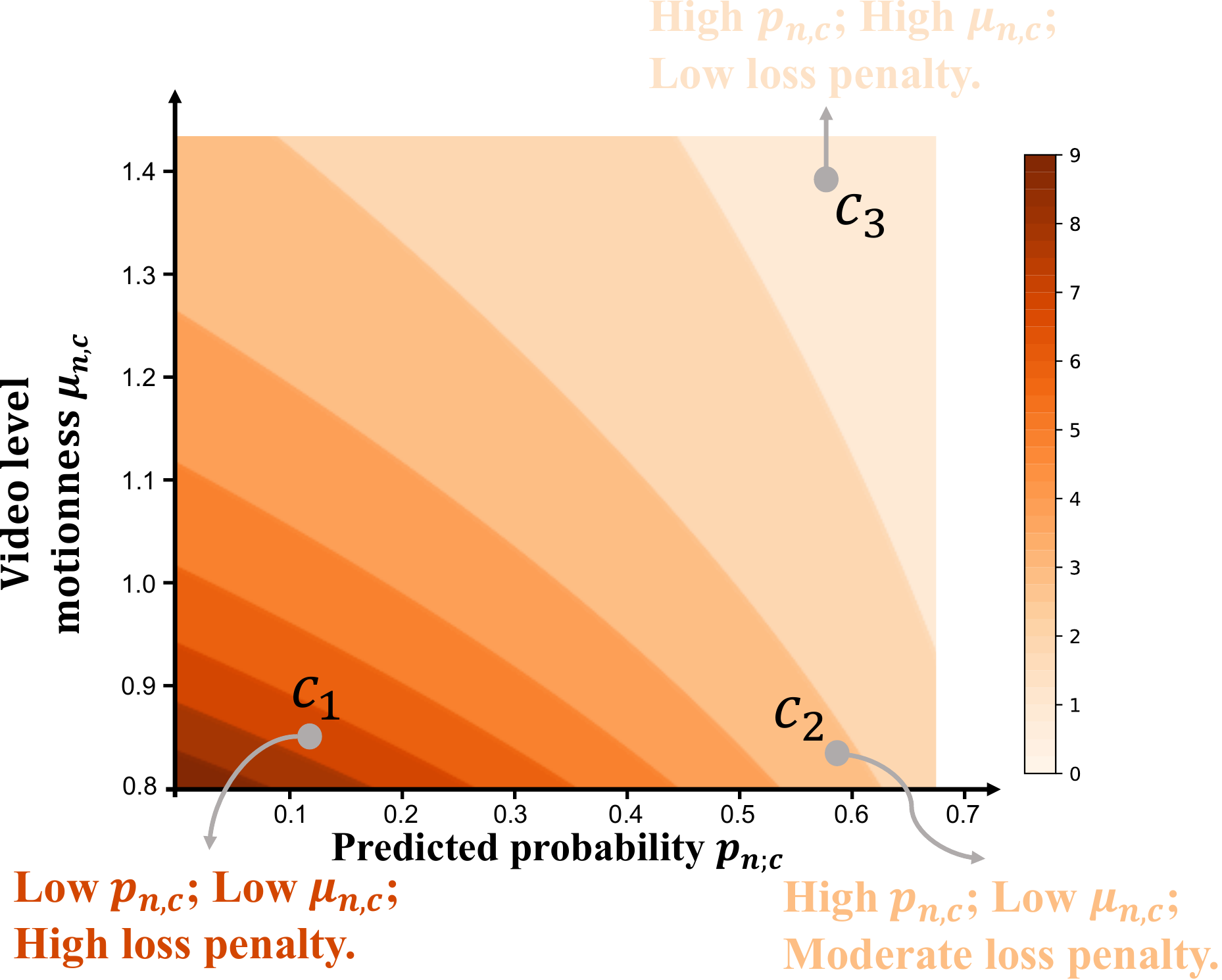}
	\end{center}
	\caption{The loss surface of our motion-guided loss. We select one action category $c$ for video $V_n$ with ground truth label 1. $p_{n;c}$ denotes the predicted probability, the same as defined in Eq.~\eqref{equation:action_loss}. $\mu_{n;c}$ denotes the video-level motionness score for the selected category $c$.}
	\label{fig:LossSuface}
\end{figure}

We interpret this formulation of Eq.~\eqref{equation:guide_loss} term by term intuitively. The first term aims to highlight the snippets with both high $p_{n;c}$ and high motionness ($\mu_{n;c}$) while the second term (\(-\log {{\mu_{n;c}}^{2}}\)) serves as a regularization term to prevent motionness from dominating the training process. For clarity, we further plot the loss surface of $\mathcal{L}_{g}$ in Fig.~\ref{fig:LossSuface}. Specifically, we randomly select one action category $c$ for video $V_n$ which has the ground truth label 1. Given the network and the trained checkpoint, we manually shift the predicted probability $p_{n;c}$ and the mean motionness score $\mu_{n;c}$ to compute the $\mathcal{L}_{g}$ value according to Eq.~\eqref{equation:guide_loss}. As shown in Fig.~\ref{fig:LossSuface}, our motion-guided loss is closely related to both the predicted probability and the video-level motionness score. To be specific, the heaviest loss penalty is applied when both the $p_{n;c}$ and $\mu_{n;c}$ are low ($c_1$ in Fig.~\ref{fig:LossSuface}). For samples with high $p_{n;c}$ yet low $\mu_{n;c}$ ($c_2$ in Fig.~\ref{fig:LossSuface}), they deserve a moderate loss penalty because of their low certainty to be action-of-interest. Samples with both high $p_{n;c}$ and $\mu_{n;c}$ ($c_3$ in Fig.~\ref{fig:LossSuface}) are given the slightest punishment.

\subsection{Inference}  \label{sec:infer}
During testing, given an input video, we first generate its actionness scores, aggregate top-$k$ scores, and threshold it with $\theta_c$ to get video-level classification results. Then for the predicted action classes, we threshold the corresponding actionness scores with threshold $\theta_a$ to select candidate snippets. Finally, we group continuous snippets into proposals and use NMS (Non-Maximum Suppression) to remove duplicates.

\section{Experiments}\label{expes}
In this section, we conducted extensive experiments to demonstrate the effectiveness of our DMP-Net. Comprehensive ablation studies on the THUMOS'14 testing set were also performed to provide more insights into each component. 

	\begin{table*}[t]
		\begin{center}
			\caption{Performance comparisons with state-of-the-art fully-supervised and weakly-supervised TAL methods on THUMOS'14 dataset. UNT and I3D are abbreviations for the UntrimmedNet feature and I3D feature, respectively.}
			\label{table:thumos_comp}
			\resizebox{\linewidth}{!}{
				\begin{tabular}{cccccccccccc}
					\toprule
					\multirow{2.5}*{\textbf{\tabincell{c}{Supervision\\(Feature)}}} & \multirow{2.5}*{\textbf{Method}} & \multirow{2.5}*{\textbf{Venue}} & \multicolumn{9}{c}{\textbf{mAP@IoU (\%)}}\\
					\cmidrule(lr){4-12}
					& & & 0.1 & 0.2 & 0.3 & 0.4 & 0.5 & 0.6 & 0.7 & 0.8 & 0.9 \\
					\midrule
					\multirow{4}*{\tabincell{c}{Full\\(--)}} & S-CNN~\cite{shou2016temporal} & \emph{CVPR 2016}  & 47.7 & 43.5 & 36.3 & 28.7 & 19.0 & 10.3 & 5.3 & -- & -- \\
					& SSN~\cite{zhao2017temporal}  & \emph{ICCV 2017} & 66.0 & 59.4 & 51.9 & 41.0 & 29.8 & -- & -- & -- & -- \\
					& P-GCN~\cite{zeng2019graph} & \emph{ICCV 2019} & 69.5 & 67.8 & 63.6 & 57.8 & 49.1 & -- & -- & -- & -- \\
					& G-TAD~\cite{xu2020g} & \emph{CVPR 2020} & -- & -- & \textbf{66.4} & \textbf{60.4} & \textbf{51.6} & \textbf{37.6} & \textbf{22.9} & -- & -- \\
					\midrule
					\midrule
					\multirow{2}*{\tabincell{c}{Weak\\(--)}} & Hide-and-Seek~\cite{singh2017hide} & \emph{ICCV 2017} & 36.4 & 27.8 & 19.5 & 12.7 & 6.8 & -- & -- & -- & -- \\
					& UntrimmedNet~\cite{wang2017untrimmednets} & \emph{CVPR 2017} & 44.4 & 37.7 & 28.2 & 21.1 & 13.7 & -- & -- & -- & -- \\
					\hdashline
					\multirow{3}*{\tabincell{c}{Weak\\(UNT)}} & AutoLoc~\cite{shou2018autoloc} & \emph{ECCV 2018} & -- & -- & 35.8 & 29.0 & 21.2 & 13.4 & 5.8 & -- & -- \\
					& CleanNet~\cite{liu2019weakly} & \emph{ICCV 2019} & -- & -- & 37.0 & 30.9 & 23.9 & 13.9 & 7.1 & -- & -- \\
					& Bas-Net~\cite{lee2020background} & \emph{AAAI 2020} & -- & -- & 42.8 & 34.7 & 25.1 & 17.1 & 9.3 & -- & -- \\
					\hdashline
					\multirow{8}*{\tabincell{c}{Weak\\(I3D)}} & STPN~\cite{nguyen2018weakly} & \emph{CVPR 2018} & 52.0 & 44.7 & 35.5 & 25.8 & 16.9 & 9.9 & 4.3 & 1.2 & 0.1 \\
					& W-TALC~\cite{paul2018w} & \emph{ECCV 2018} & 55.2 & 49.6 & 40.1 & 31.1 & 22.8 & -- & 7.6 & -- & -- \\
					& Liu \emph{et al.}~\cite{liu2019completeness}  & \emph{CVPR 2019} & 57.4 & 50.8 & 41.2 & 32.1 & 23.1 & 15.0 & 7.0 & -- & -- \\
					& Nguyen \emph{et al.}~\cite{nguyen2019weakly} & \emph{ICCV 2019} & 60.4 & 56.0 & 46.6 & 37.5 & 26.8 & 17.6 & 9.0 & 3.3 & 0.4 \\
					& BaS-Net~\cite{lee2020background} & \emph{AAAI 2020} & 58.2 & 52.3 & 44.6 & 36.0 & 27.0 & 18.6 & 10.4 & 3.9 & 0.5 \\
					& DGAM~\cite{shi2020weakly} & \emph{CVPR 2020} & 60.0 & 54.2 & 46.8 & 38.2 & 28.8 & 19.8 & 11.4 & 3.6 & 0.4 \\
					& UM-Net~\cite{lee2021weakly} & \emph{AAAI 2021} & 67.5 & 61.2 & 52.3 &  43.4 & 33.7 & 22.9 & 12.1 & -- & -- \\
					& SF-Net~\cite{ma2020sf} & ECCV 2020 & -- & -- & 52.8 & 42.2 & 30.5 & 20.6 & 12.0 & -- & --\\
					& CoLA~\cite{zhang2021cola} & \emph{CVPR 2021} & 66.2 & 59.5 & 51.5 & 41.9 & 32.2 & 22.0 & 13.1 & -- & -- \\
					& \textbf{DMP-Net (Ours)} & -- & \textbf{68.1} & \textbf{61.7} & \textbf{52.9} & \textbf{44.0} & \textbf{34.2} & \textbf{23.5} & \textbf{13.1} & \textbf{4.7} & \textbf{0.6} \\
					\bottomrule
				\end{tabular}
			}
		\end{center}
	\end{table*}

\subsection{Datasets}\label{sec:4_1}
We evaluated DMP-Net on three prevailing and challenging WSTAL benchmarks: \emph{THUMOS'14}, \emph{ActivityNet v1.2} and \emph{ActivityNet v1.3}. Although frame-level labels are provided in these datasets, we only use the video-level category labels for network training.

\textbf{THUMOS'14}~\cite{idrees2017thumos}. It is a widely-adopted benchmark for WSTAL, which consists of untrimmed videos from 20 sports categories. The training, validation, and test sets contain 13,320, 1,010, and 1,574 videos, respectively. It is very challenging since each video in this dataset may contain multiple action instances, with an average of 1.12 classes per video. Following the common setting in~\cite{lee2020background,shi2020weakly}, we used the validation set (200 videos) for training and the test set (213 videos) for testing. 

\textbf{ActivityNet}~\cite{caba2015activitynet}. It is a large-scale benchmark for WSTAL, which contains two versions: ActivityNet v1.3 and ActivityNet v1.2. ActivityNet v1.3 consists of 19,994 untrimmed videos from 200 classes. The training, validation, and test splits are divided by the ratio of 2:1:1. ActivityNet v1.2 is a subset of ActivityNet v1.3, which covers 100 action categories with 4,819 training, 2,383 validation and 2,480 test videos. Each video has an average of 1.65 action instances. Following the common practice~\cite{wang2017untrimmednets,shou2018autoloc}, we trained the models on the training set and evaluated them on the validation set.

\subsection{Implementation Details}

\textbf{Evaluation Metrics.} Following the standard evaluation protocol, we reported mean Average Precision (mAP) values under different Intersection over Union (IoU) thresholds. For THUMOS'14, the IoU thresholds are set to $\{0.1, 0.2, ... , 0.9\}$. For ActivityNet, the thresholds are chosen from $\{0.5,0.75,0.95\}$, and we also report the average mAP of all the IoU thresholds between $0.5$ and $0.95$ with the step of $0.05$. The evaluations of both two versions are calculated by using the official codes\footnote{\url{https://github.com/activitynet/ActivityNet/}}.

\textbf{Feature Extractor.} We used the I3D network~\cite{carreira2017quo} pre-trained on Kinetics~\cite{carreira2017quo} as our feature extractor. Note that the I3D feature extractor is not fine-tuned for fair comparisons. We applied the TVL1~\cite{perez2013tv} algorithm to extract optical flow in advance. Video snippets were sampled every 16 frames and the feature dimension for each extracted snippet was 1,024. 

\textbf{Training Details.} The sampling number $T$ was set as 750 for THUMOS'14 and 50 for ActivityNet, respectively. Following~\cite{nguyen2018weakly,lee2020background}, we performed stratified random perturbation on the segments sampled for data augmentation and used the uniform sampling strategy during the test. All hyper-parameters were determined by grid search: GCN layer number $K=2$, selection ratio $r=8$. We set the embedding dimension in GCN as 1024. For all datasets, positional threshold $\theta_{pos}$ and similarity threshold $\gamma$ were set to 0.1 and 0.6, respectively. We used the Adam optimizer with a learning rate of 1e-4. We trained for a total of 6k epochs with a batch size of 32 for THUMOS'14 dataset and 8k epochs with a batch size of 128 for ActivityNet dataset. Experiments were conducted on a single V100 GPU.

\textbf{Inference Details.} We set $\theta_c$ to 0.2 and 0.1 for the THUMOS’14 and ActivityNet datasets, respectively. For proposal generation, we used multiple thresholds with $\theta_a$ set to [0:0.25:0.025] for THUMOS'14 and [0:0.15:0.015] for ActivityNet. Non-Maximum Suppression (NMS) was applied with IoU threshold 0.7.
\subsection{Comparisons with State-of-the-Arts Methods} \label{sec:4_2}
We compare our DMP-Net with the state-of-the-art WSTAL methods on three challenging datasets:

\textbf{THUMOS'14.}
We compared our DMP-Net with the state-of-the-art fully-supervised and weakly-supervised TAL approaches on the THUMOS'14 testing set. As shown in TABLE~\ref{table:thumos_comp}, DMP-Net consistently outperforms previous weakly-supervised methods at all IoU thresholds, with mAP@0.5 reaching 34.2\%. Moreover, DMP-Net even surpasses several strong fully-supervised methods (\emph{e.g.}, S-CNN~\cite{shou2016temporal} and SSN~\cite{zhao2017temporal}) with less supervision.
	
\textbf{ActivityNet v1.2.}
The comparison results on the ActivityNet v1.2 validation set are summarized in TABLE~\ref{table:anet12_comp}. Our DMP-Net achieves state-of-the-art 26.1\% average mAP among all weakly supervised TAL methods. Besides, the performance of DMP-Net is still competitive when compared with the fully-supervised method SSN~\cite{zhao2017temporal}, \ie, it even surpasses SSN at mAP@IoU 0.5 (41.9\% \emph{v.s.} 41.3\%).

\textbf{ActivityNet v1.3.}
We conducted experiments on the ActivityNet v1.3 validation set and the comparison results are reported in TABLE~\ref{table:anet13_comp}. On this larger version dataset, our DMP-Net also achieves competitive performance with average mAP reaching 23.9\%. The consistent superior results on both versions of the ActivityNet dataset signify the effectiveness of DMP-Net. 

\begin{table}[t]
		\begin{center}
			\caption{Comparison results on ActivityNet v1.2 dataset. The AVG column shows the averaged mAP under the thresholds 0.5:0.05:0.95. UNT and I3D are abbreviations for UntrimmedNet feature and I3D feature, respectively.}
			\label{table:anet12_comp}
			\resizebox{\linewidth}{!}{
				\begin{tabular}{ccccccc}
					\toprule
					\multirow{2.5}*{\textbf{Sup.}} & \multirow{2.5}*{\textbf{Method}} & \multicolumn{4}{c}{\textbf{mAP@IoU (\%)}}\\
					\cmidrule(lr){3-7}
					& & 0.5 & 0.75 & 0.95 & AVG \\
					\midrule
					Full & SSN~\cite{zhao2017temporal} & 41.3 & 27.0 & 6.1 & 26.6\\
					\midrule
					Weak & UntrimmedNet~\cite{wang2017untrimmednets} & 7.4 & 3.2 & 0.7 & 3.6\\
					(UNT)& AutoLoc~\cite{shou2018autoloc} & 27.3 & 15.1 & 3.3 & 16.0\\
					\hdashline
					\multirow{7}*{\tabincell{c}{Weak\\(I3D)}} & W-TALC~\cite{paul2018w} & 37.0 & 12.7 & 1.5 & 18.0\\
					& TSM~\cite{yu2019temporal} & 28.3 & 17.0 & 3.5 & 17.1\\
					& CleanNet~\cite{liu2019weakly} & 37.1 & 20.3 & 5.0 & 21.6\\
					& Liu \emph{et al.}~\cite{liu2019completeness} & 36.8 & 22.0 & 5.6 & 22.4\\
					& BaS-Net~\cite{lee2020background} & 38.5 & 24.2 & 5.6 & 24.3\\
					& DGAM~\cite{shi2020weakly} & 41.0 & 23.5 & 5.3 & 24.4\\
					& UM-Net~\cite{lee2021weakly} & 41.2 & 25.6 & 6.0 & 25.9\\
					& SF-Net~\cite{ma2020sf} & 37.8 & -- & -- & 22.8 \\ 
					& \textbf{DMP-Net (Ours)} & \textbf{41.9} & \textbf{27.0} & \textbf{6.2} & \textbf{26.1}\\
					\bottomrule
				\end{tabular}
			}
		\end{center}
\end{table}

\begin{table}[t]
		\begin{center}
			\caption{Comparison results on ActivityNet v1.3 dataset. The AVG column shows the averaged mAP under the thresholds 0.5:0.05:0.95. All listed methods use the I3D feature.}
			\label{table:anet13_comp}
			\resizebox{\linewidth}{!}{
				\begin{tabular}{ccccccc}
					\toprule
					\multirow{2.5}*{\textbf{Sup.}} & \multirow{2.5}*{\textbf{Method}} & \multicolumn{4}{c}{\textbf{mAP@IoU (\%)}}\\
					\cmidrule(lr){3-7}
					& & 0.5 & 0.75 & 0.95 & AVG \\
					\midrule
					\multirow{6}*{\tabincell{c}{Weak}} & STPN~\cite{nguyen2018weakly} & 29.3 & 16.9 & 2.6 & 16.3\\
					& TSM~\cite{yu2019temporal} & 30.3 & 19.0 & 4.5 & -\\
					& BaS-Net~\cite{lee2020background} & 34.5 & 22.5 & 4.9 & 22.2\\
					& Liu \emph{et al.}~\cite{liu2019completeness} & 34.0 & 20.9 & \textbf{5.7} & 21.2 \\
					& A2CL-PT~\cite{min2020adversarial} & 36.8 & 22.0 & 5.2 & 22.5 \\
					& \textbf{DMP-Net (Ours)} & \textbf{37.9} & \textbf{23.5} & 5.3 & \textbf{23.9}\\
					\bottomrule
				\end{tabular}
			}
		\end{center}
\end{table}
	
\subsection{Analysis of Deep Motion Prior}\label{sec:4_3_1}\label{subsec:ana_dmp}
In our paper, we emphasize the importance of the optical flow modality and build a deep motion prior to guide the training of the whole network. Here we design some pilot experiments to verify the necessity of our motivation.

\subsubsection{\textbf{Optical flow v.s. RGB modalities}} To intuitively compare the two modalities, we conducted comparative experiments based on each single modality. Specifically, we selected two representative WSTAL methods: BaS-Net~\cite{lee2020background} and TSCN~\cite{zhai2020two}, which belong to early-fusion and late-fusion types, respectively. UM-Net takes the concatenation of the RGB and the optical flow modalities as input. TSCN is trained based on the two features separately and then the outputs of each branch are summed up. For both methods, we replaced the network input with only a single modality feature and retrained the network. These two model variants are called flow-based and RGB-based approaches while the original model is called the two-stream-based approach. All results are reported in TABLE~\ref{table:rgbvsflow}.

\textit{Quantitative Results.} The results in TABLE~\ref{table:rgbvsflow} show that the performance of the flow-based approaches is slightly lower than the two-stream-based approaches, but far exceeds that of the RGB-based ones. For example, on mAP@IoU 0.5 of BaS-Net, the flow-based approach decreases by only 0.6\% over the official two-stream approach, while the RGB-based one has a dramatic 8.6\% performance drop. The results show that the RGB modality is less sensitive to actions~\cite{zhai2020two} and further verify our hypothesis that the optical flow modality is more informative than the RGB modality.

\begin{table}[t]
		\begin{center}
			\caption{Pilot experiments on THUMOS'14 of training with different modalities. ``R \& F" feature denotes the two-stream based approach, \ie, it concatenates RGB and Flow modalities.}
			\label{table:rgbvsflow}
			\resizebox{0.9\linewidth}{!}{
				\begin{tabular}{cccccc}
					\toprule
					\multirow{2.5}*{\textbf{Method}} & \multirow{2.5}*{\textbf{Feature}} & \multicolumn{4}{c}{\textbf{mAP@IoU (\%)}}\\
					\cmidrule(lr){3-6}
					& & 0.4 & 0.5 & 0.6 & 0.7 \\
					\midrule
					\multirow{3}*{\tabincell{c}{BaS-Net~\cite{lee2020background}}} & RGB & 21.5 & 18.4 & 13.8 & 8.4\\
					& Flow & 35.6 & 26.4 & 17.9 & 9.9\\
					& R\&F & \textbf{36.0} & \textbf{27.0} & \textbf{18.6} & \textbf{10.4} \\
					\midrule
					\multirow{3}*{\tabincell{c}{TSCN~\cite{zhai2020two}}} & RGB & 22.3 & 13.2 & 10.7 & 5.5\\
					& Flow & 34.8 & 24.6 & 17.3 & 8.7\\
					& R\&F & \textbf{37.7} & \textbf{28.7} & \textbf{19.4} & \textbf{10.2} \\
					\bottomrule
				\end{tabular}
			}
		\end{center}
\end{table}

\subsubsection{\textbf{Learning motion prior from different modalities}} In our method, we feed the optical flow modality to the guidance branch to generate the deep motion prior. Here we conducted comparison experiments that take the RGB or two-stream modality as input to the guidance branch. Besides, for more intuitive understanding, we also compute KL divergence values between the ground truth and the guidance sequence. Results are reported in TABLE~\ref{table:vsCEMGAblation}.

\textit{Quantitative Results.} The results in TABLE~\ref{table:vsCEMGAblation} show that both the RGB guided and the two-stream guided variants lead to performance degradation. For example, mAP@IoU 0.5 of the RGB guided variant is only 22.2\%, much lower than 34.2\% of the optical flow guided version. Besides, the motionness sequence generated by $\mathcal{L}_{g, Flow}$ has the most correlated distributions with ground truth according to the KL divergence values. Thus, we conclude that the RGB modality is not suitable for our motionness modeling and can not guide the network training.

\begin{table}[t]
		\begin{center}
			\caption{Ablation studies on THUMOS'14 of different loss functions for DMP-Net. $\mathcal{L}_{a}$ denotes the traditional XE loss. $\mathcal{L}_{g, Flow}$, $\mathcal{L}_{g, RGB}$ and $\mathcal{L}_{g, R \& F}$ represent the motion-guided (ours), RGB-guided, and two-stream guided losses, respectively. $\mathcal{L}_{g, Flow'}$ denotes the motion-guided loss variant without the regularization term. KL denotes the KL divergence between the ground truth and the guidance sequence.}
			\label{table:vsCEMGAblation}
				\begin{tabular}{cccccc}
					\toprule
					\multirow{2.5}*{\textbf{Loss}}  & \multicolumn{4}{c}{\textbf{mAP@IoU (\%)}} & \multirow{2.5}*{\textbf{KL}} \\
					\cmidrule(lr){2-5}
					&  0.4 & 0.5 & 0.6 & 0.7  \\
					\midrule
					$\mathcal{L}_{g, Flow}$ & \textbf{44.0} & \textbf{34.2} & \textbf{23.5} & \textbf{13.1} & \textbf{0.004} \\
					$\mathcal{L}_{g, RGB}$ & 31.9 & 22.2 & 13.5 & 4.6 & 0.087 \\
					$\mathcal{L}_{g, R \& F}$ & 40.5 & 28.3 & 20.8 & 11.0 & 0.023 \\
					\hdashline
					$\mathcal{L}_{a}$ & 33.1 & 23.8 & 15.2 & 7.9  & -- \\
					\hdashline
					$\mathcal{L}_{g, Flow'}$ & 40.7 & 30.4 & 19.7 & 11.4 & 0.016  \\
					\bottomrule
				\end{tabular}
		\end{center}
\end{table}


\subsection{Ablation Studies on Motion-guided Loss}\label{sec:4_3_2}\label{subsec:ana_loss}
Leveraging the deep motion prior, we propose a motion-guided loss to modulate the network training. Here more ablation studies are conducted to give in-depth analysis from both quantitative and qualitative perspectives.

\subsubsection{\textbf{Motion-guided Loss v.s. XE Loss}} We conducted comparison experiments supervised by the traditional cross-entropy loss $\mathcal{L}_a$ (cf. Sec.~\ref{sec:loss}). All other experimental settings remained unchanged except for the training loss.

\textit{Quantitative Results.} Results in TABLE~\ref{table:vsCEMGAblation} demonstrate that replacing $\mathcal{L}_{g, Flow}$ with $\mathcal{L}_a$ results in significant performance drops (\eg, 10.4\% at mAP@0.5), which proves that motionness modeling is essential and the motion-guided loss can efficiently guide network training with this deep motion prior. 

\begin{figure}[t]
	\begin{center}
		\includegraphics[width=0.7\linewidth]{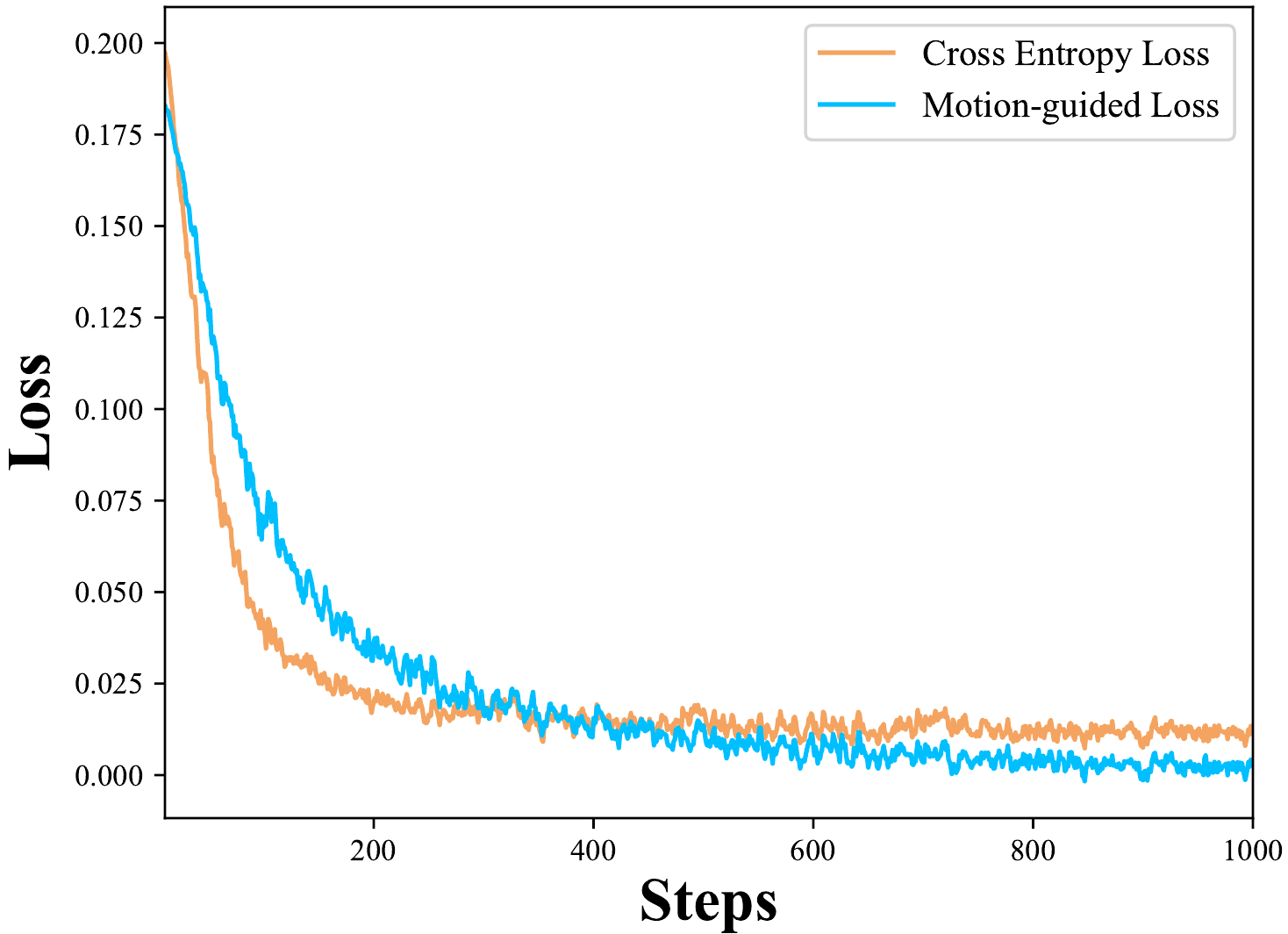}
	\end{center}
	\caption{Loss curves of XE loss (orange) and motion-guided loss (blue) for DMP-Net on THUMOS'14.}
	\label{fig:CEvsMG}
\end{figure}

\textit{Qualitative Results.} We presented the loss curves of the cross-entropy loss $\mathcal{L}_a$ and motion-guided loss $\mathcal{L}_{g, Flow}$ in Fig.~\ref{fig:CEvsMG}, which shows that our motion-guided loss leads to a more gentle descent process yet converges to a lower loss value. 

\begin{figure}[t]
		\begin{center}
			\includegraphics[width=\linewidth]{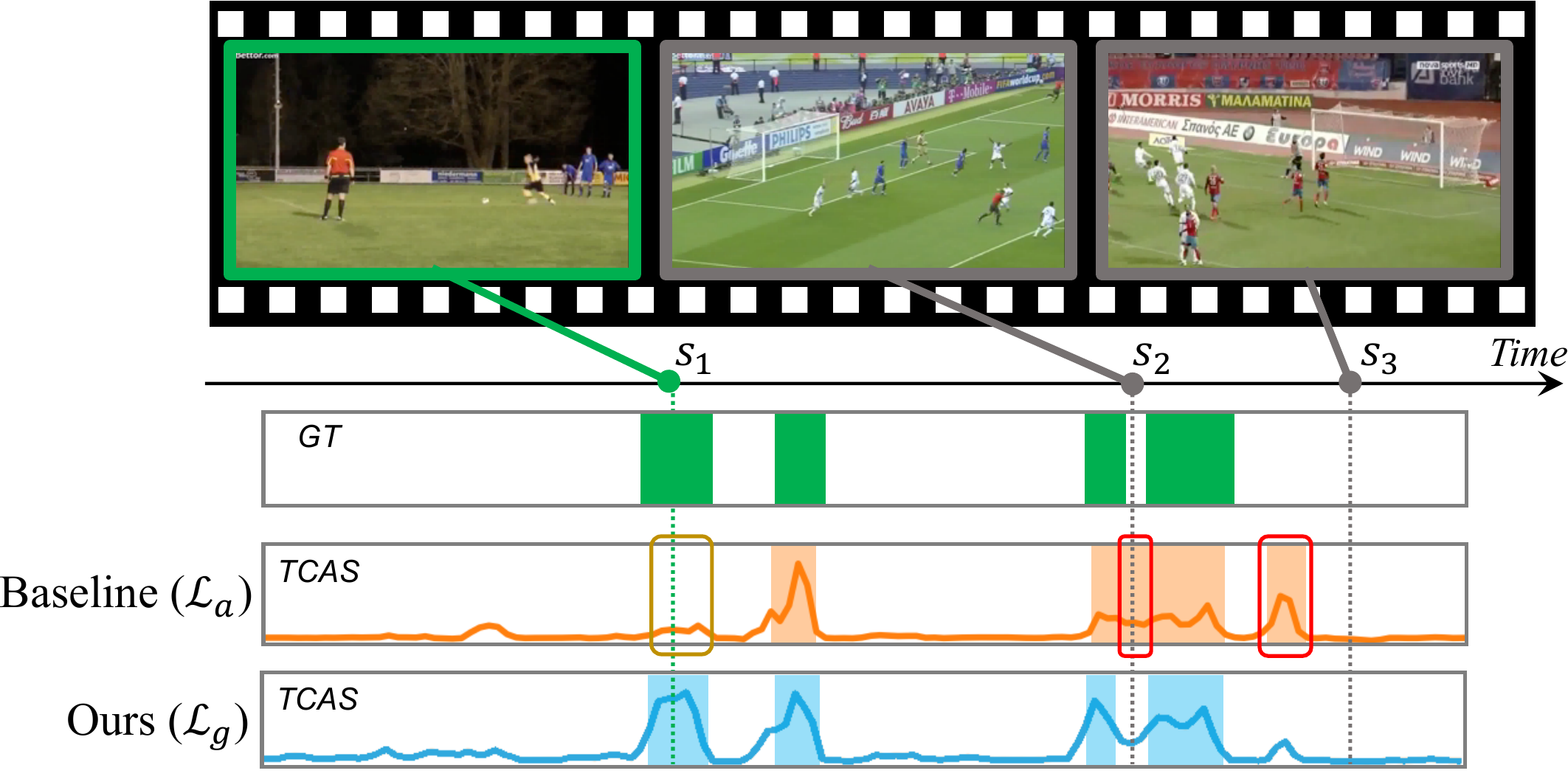}
		\end{center}
		\caption{An example of \texttt{SoccerPenalty} video from THUMOS'14. GT denotes ground truth action instances. TCAS denotes the TCAS distributions for category \texttt{SoccerPenalty} trained with $\mathcal{L}_{g}$ and $\mathcal{L}_{a}$. Prediction results are obtained by thresholding and visualized by corresponding masks.}
		\label{fig:ab_motion}
	\end{figure}

Besides, we randomly selected one video from the THUMOS'14 testing set and visualize TCAS results of well-trained models using $\mathcal{L}_a$ and $\mathcal{L}_g$, respectively\footnote{More qualitative results and video demos are available in the supplementary materials. \label{fn:supple}}. In the \texttt{SoccerPenalty} video of Fig~\ref{fig:ab_motion}, the $\mathcal{L}_{a}$ equipped model fails to locate the 1st scene (\ie, \textit{$s_1$}) maybe because the football goal does not appear in the camera shot. Besides, the mass celebration after scoring (\ie, $s_3$) is also mistaken as the action. Our DMP-Net effectively filters out these error-prone clips.

\subsubsection{\textbf{Ablations of Motion-guided Loss components}} We apply a regularization term in the motion-guided loss computation, which prevents motionness from dominating the network training. The ablation studies for this term are presented in TABLE.~\ref{table:vsCEMGAblation}.

\textit{Quantitative Results.} From TABLE.~\ref{table:vsCEMGAblation}, we find that the loss variant without the regularization term has the inferior performance to the full version. For example, on mAP@IoU 0.5, the performance drops by 3.8\% (19.7\% \emph{v.s.} 23.5\%), which demonstrates that our regularization term is beneficial to the network training.

\subsection{Analysis of Motion Graph}\label{sec:4_3_3}\label{subsec:ana_graph}
	\begin{figure}[t]
		\begin{center}
			\includegraphics[width=0.9\linewidth]{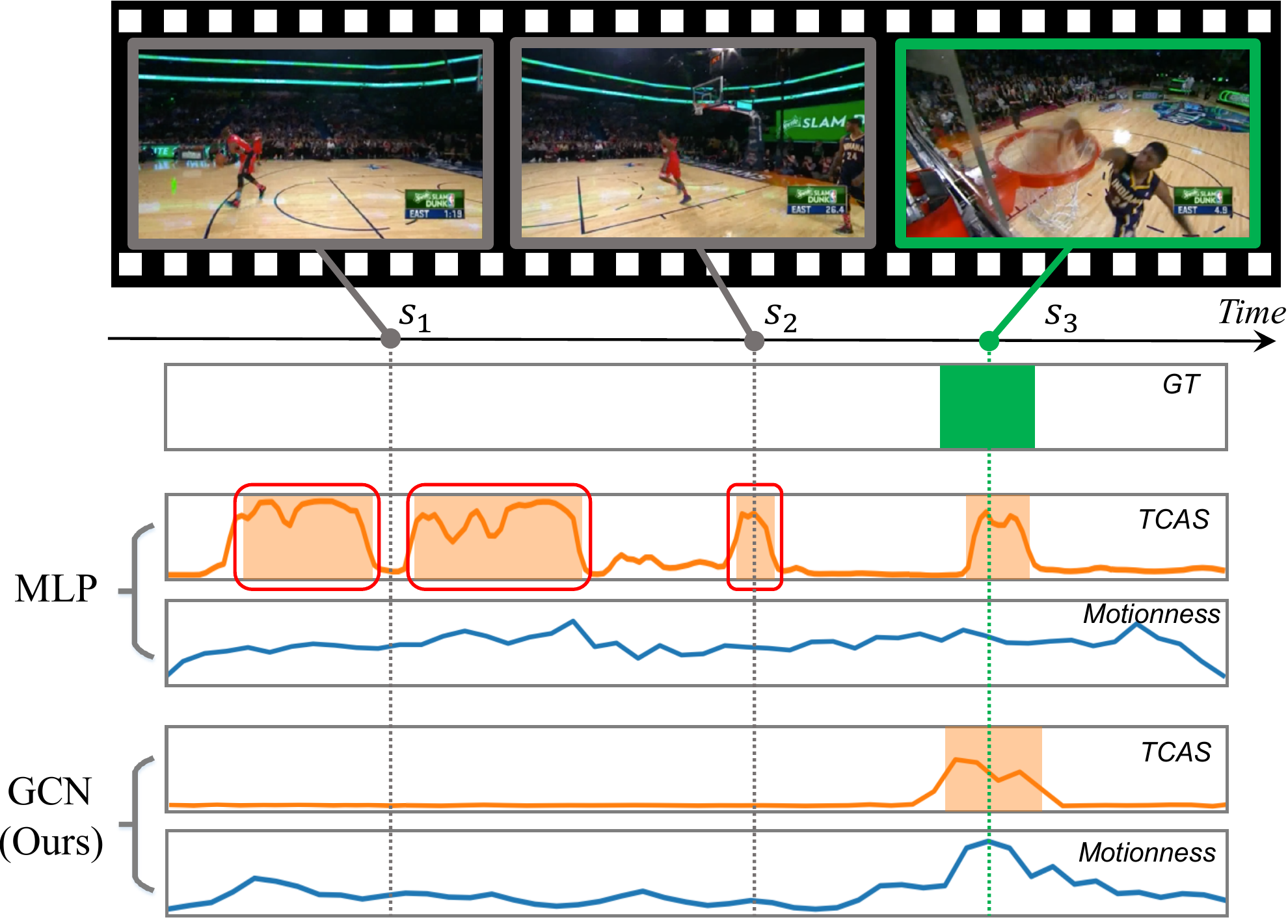}
		\end{center}
		\caption{An example of \texttt{BasketballDunk} video from THUMOS'14. GT denotes ground truth action instances. TCAS for category \texttt{BasketballDunk} and motionness distributions using MLP and GCN (Ours) are presented. Prediction results are obtained by thresholding and visualized by corresponding masks.}
		\label{fig:ab_GCN}
	\end{figure}

\begin{table}[t]
		\begin{center}
			\caption{Ablation studies on the motion graph. MLP denotes using MLP to model the snippet-wise correlations. Dense represents building the motion graph in a fully connected manner. Sparse is our proposed motion graph with positional edges $\mathcal{E}_{pos}$ and semantic edges $\mathcal{E}_{smt}$.}
			\label{table:GraphAblation}
			\resizebox{0.85\linewidth}{!}{
				\begin{tabular}{cccccc}
					\toprule
					\multicolumn{2}{c}{\multirow{2.5}*{\textbf{Mode}}} &  \multicolumn{4}{c}{\textbf{mAP@IoU (\%)}}\\
					\cmidrule(lr){3-6}
					& & 0.4 & 0.5 & 0.6 & 0.7 \\
					\midrule
					\multicolumn{2}{c}{MLP} & 39.1 & 30.2 & 19.2 & 10.4 \\
					\hdashline
					\multicolumn{2}{c}{Dense} & 41.0 & 32.8 & 21.5 & 11.5 \\
					\hdashline
					\multirow{3}*{\tabincell{c}{Sparse}} & w/ all edges & \textbf{44.0} & \textbf{34.2} & \textbf{23.5} & \textbf{13.1}\\
					& w/o $\mathcal{E}_{pos}$ &  42.7 & 33.6 & 22.7 & 12.4\\
					& w/o $\mathcal{E}_{smt}$  & 41.5 & 33.3 & 21.6 & 11.5\\
					\bottomrule
				\end{tabular}
			}
		\end{center}
\end{table}
To obtain the context-dependent motion prior, we construct a motion graph with positional edges and semantic edges. Here we present ablation studies of the graph components and compare our sparse graph to the fully-connected one.

\subsubsection{\textbf{The necessity of modeling relationship between nodes}} As illustrated in Sec.~\ref{subsec: motionness}, we introduced graph convolution layers to help enable information dissemination among snippets. To demonstrate its efficacy, we implemented the motionness modeling with a 2-layer MultiLayer-Perceptron (MLP) for comparison. Specifically, we discarded the adjacent matrix in Eq.~\eqref{equ:gcn_motion}, namely using \(\bm{X}^{k}=\bm{X}^{k-1}\bm{W}^k\) for each layer updating, where $\bm{W}^k$ are learnable parameters.

\textit{Quantitative Results.} As shown in TABLE~\ref{table:GraphAblation}, GCN leads to better performance at all IoUs, which justifies its superiority in the message passing among snippets.

\textit{Qualitative Results.} We also chose one video and visualized the TCAS and motionness distributions for two variants using GCN and MLP, respectively\ref{fn:supple}. The results in Fig.~\ref{fig:ab_GCN} demonstrate that GCN helps capture action-of-interest and get rid of the distracting actions, \eg, player dribbling is mistakenly highlighted by MLP, and the corresponding TCAS is dominated by such misleading background actions. 

\subsubsection{\textbf{Ablations studies on positional and semantic edges}} Positional edges and semantic edges are designed to capture surrounding and remote yet semantically related nodes, respectively. We conducted ablation studies on the graph components, and all results are reported in Table~\ref{table:GraphAblation}.

\textit{Quantitative Results.} As shown in Table~\ref{table:GraphAblation}, removing any type of edge leads to the remarkable performance drop, especially for the semantic edges. For example, mAP@IoU 0.6 drops by 1.9\% in the absence of $\mathcal{E}_{smt}$. Thus both kinds of edges play a very important role in relationship modeling. 

\subsubsection{\textbf{Sparse graph v.s. Dense-connected graph}} An intuitive way to build the motion graph is to connect all nodes in a \emph{dense} manner, \ie, edges are built between all possible node pairs. Specifically, following Eq.~\eqref{equ:adj_def}, we computed the adjacent matrix as $\bm{G}_{i j}=\frac{(\mathbf{W}_{1}\bm{m}_i)^{\top} (\mathbf{W}_{2}\bm{m}_j)}{\sum_{j=1}^{T} (\mathbf{W}_{1}\bm{m}_i)^{\top} (\mathbf{W}_{2}\bm{m}_j)}$, where $\mathbf{W}_{1}$ and $\mathbf{W}_{2}$ are learnable parameters and $T$ is the number of temporal nodes. Obviously, this fully-connected approach is more computation intensive and requires more storage resources.

\textit{Quantitative Results.} As shown in TABLE~\ref{table:GraphAblation}, the performance of the fully-connected graph is behind our sparse graph (\ie, w/ all edges). For example, the performance drops by 1.4\% at mAP@IoU 0.5 (34.2\% \emph{v.s.} 32.8\%).

\begin{figure}[t]
	\begin{center}
		\includegraphics[width=\linewidth]{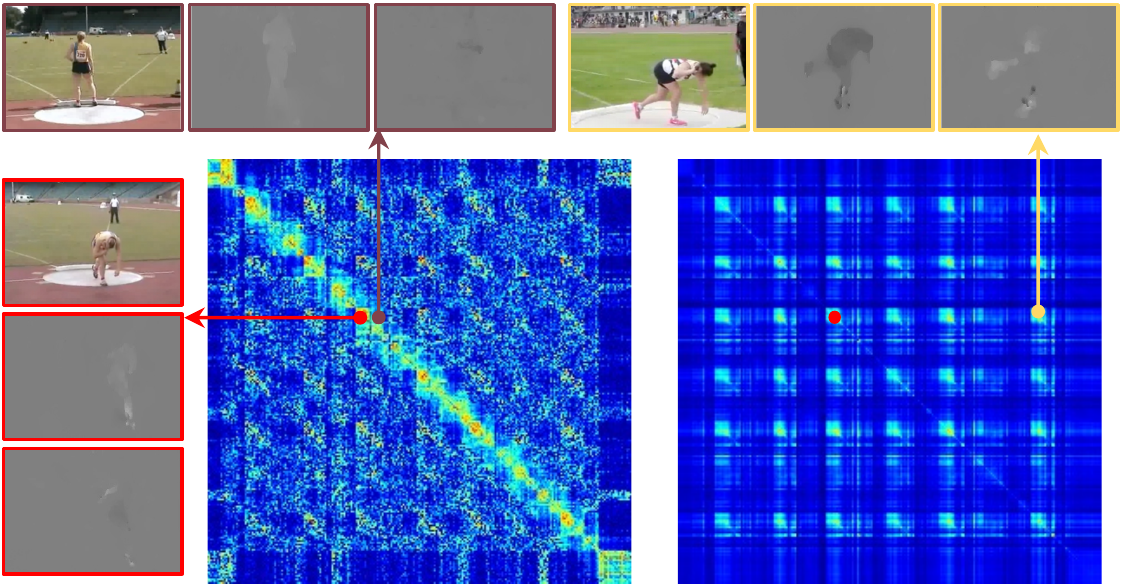}
	\end{center}
	\caption{Visualizations of the adjacent matrix for an example video of \texttt{Shotput}. \textbf{Left:} dense connected graph; \textbf{Right:} our sparse connection. We select one temporal node marked in \textcolor{red}{\textbf{red}}. The nodes with high correlation scores are presented in \textcolor[RGB]{129,65,75}{\textbf{brown}} and \textcolor[RGB]{254,217,102}{\textbf{yellow}} for dense and sparse graphs, respectively.}
	\label{fig:adjMatix}
\end{figure}

\textit{Qualitative Results.} To reveal the rationale behind this, we selected one video from the THUMOS'14 testing set and visualized the adjacent matrix of both the dense connected and our sparse graph in Fig.~\ref{fig:adjMatix}. We can easily observe that the high weight of the adjacency matrix of the fully-connected graph is mainly concentrated in the diagonal area, \ie, each node is more relevant to its surrounding nodes. This may be due to the \textit{slowness} prior~\cite{zhang2012slow} of video data, where short-term features change slowly in a local window. Therefore, the adjacent weight is focused on the positional adjacent areas while neglecting the remote yet semantic correlated snippets. In contrast, our sparse graph alleviates this problem by encouraging the semantic correlations of snippets even when they are far away. For example, in Fig.~\ref{fig:adjMatix}, the \texttt{Shotput} action conducted by two different actors is scattered in a long temporal range. Our motion graph can capture these semantic connections while the dense connection way ignores them.

\begin{table}[t]
		\begin{center}
			\caption{Model scalability on other backbones. * denotes replacing original XE loss with our motion-guided loss.}
			\label{table:moreBaseline}
			\resizebox{0.7\linewidth}{!}{
				\begin{tabular}{ccccc}
					\toprule
					\multirow{2.5}*{\textbf{Model}}  & \multicolumn{4}{c}{\textbf{mAP@IoU (\%)}}\\
					\cmidrule(lr){2-5}
					&  0.4 & 0.5 & 0.6 & 0.7  \\
					\midrule
					BaS-Net~\cite{lee2020background} & 36.0 & 27.0 & 18.6 & 10.4  \\
					BaS-Net* & \textbf{39.8} & \textbf{31.2} & \textbf{20.4} & \textbf{11.2} \\
					\hdashline
					UM-Net~\cite{lee2021weakly} & 43.4 & 33.7 & 22.9 & 12.1 \\
					UM-Net* & \textbf{44.2} & \textbf{34.5} & \textbf{23.2} & \textbf{12.8}\\
					\bottomrule
				\end{tabular}
			}
		\end{center}
\end{table}

\begin{table}[!t]
	\begin{center}
		\caption{Ablation analysis on the positional threshold $\theta_{pos}$ and similarity threshold $\gamma$.}
		\label{table:ablaParam}
		\resizebox{0.8\linewidth}{!}{
			\begin{tabular}{cccccc}
				\toprule
				\textbf{$\theta_{pos}$} & 0.01 & 0.05 & 0.1 & 0.2 & 0.3\\
				\midrule
				\textbf{mAP@0.5} & 33.5 & 33.9 & 34.2 & 33.0 & 32.8\\
				\midrule
				\midrule
				\textbf{$\gamma$} & 0.4 & 0.5 & 0.6 & 0.7 & 0.8\\
				\midrule
				\textbf{mAP@0.5} & 34.3 & 34.2 & 34.2 & 32.1 & 31.4 \\
				\bottomrule
			\end{tabular}
		}
	\end{center}
\end{table}

\subsection{Analysis of Model Scalability}\label{sec:4_3_4}\label{subsec:ana_scale}
Our proposed motion-guided loss can be easily integrated into other WSTAL frameworks as an alternative to the XE loss. Thus, we conducted more experiments to demonstrate the model scalability.

\subsubsection{\textbf{More experiments on other backbones}} To prove the general effectiveness of our motion-guided loss, in addition to the baseline backbone applied above, we selected two typical WSTAL models (BaS-Net~\cite{lee2020background} and UM-Net~\cite{lee2021weakly}) and replaced the original XE loss with our proposed motion guided loss. Specifically, for both methods, we additionally introduced our guidance branch and used it to guide the network training in the format of motion-guided loss. All the hyper-parameter settings remained the same as in our baseline. The results are reported in TABLE~\ref{table:moreBaseline}.

\textit{Quantitative Results.} As shown in TABLE~\ref{table:moreBaseline}, our motion-guided loss leads to consistent improvement for both methods. For example, in BaS-Net, mAP@IoU 0.5 is improved by 4.2\% after applying our motion-guided loss. These results demonstrate that our motion-guided loss is general and compatible with different backbones.

\subsubsection{\textbf{Analysis of the hyper-parameter sensitivity}} We conducted experiments on two important hyper-parameters including the positional threshold $\theta_{pos}$ and similarity threshold $\gamma$.  

\textit{Quantitative Results.} As shown in TABLE~\ref{table:ablaParam}, mAP@0.5 hits the peak performance when setting $\theta_{pos}=0.1$. We have noticed that too large or too small values of $\theta_{pos}$ both lead to performance degradation. This may be because a small $\theta_{pos}$ value will affect the modeling of local correlations while too large $\theta_{pos}$ value will make the network similar to the fully-connected graph, which has been demonstrated to be a sub-optimal design. For the value of $\gamma$, we see that when $\gamma$ is greater than 0.6, performance degrades rapidly. On the other hand, the cost of the number of edges decreases as the value of $\gamma$ increases. To trade-off between both the accuracy and the computation overhead comprehensively, we set $\gamma=0.6$.

\subsubsection{\textbf{Analysis of model complexity}} We conduct the model complexity analysis for both baseline and our DMP-Net. We set up three evaluation metrics, \ie, \#param, GFLOPs, and run-time. Run-time is defined as the average inference time to localize one untrimmed video.

As shown in TABLE~\ref{table:complexity}, DMP-Net introduces extra computational costs during the training process. However, considering the significant performance gains, these costs are acceptable. Besides, DMP-Net shares the same inference process as the baseline model and therefore the introduced guidance branch does not affect the reasoning speed.

\begin{table}[t]
	\caption{Model complexity analysis. Run-time is defined as the average inference time to localize one untrimmed video.}
	\label{table:complexity}
	\begin{center}
		\begin{tabular}{x{35}x{35}x{35}x{35}x{35}}
			\toprule
			 \textbf{Method}  &  \textbf{\#Params}   &   \textbf{GFLOPs} &   \textbf{run-time} & \textbf{mAP@0.5} \\
			\midrule
			Baseline & 13.27M & 29.34G & 0.048s & 23.8 \\
			DMP-Net & 16.84M & 35.53G & 0.048s & 34.2\\
			\bottomrule
		\end{tabular}
	\end{center}
\end{table}


\section{Conclusions}
In this paper, we start from summarizing the two overlooked issues in existing Weakly-Supervised Temporal Action Localization (WSTAL) methods, \ie, inadequate use of optical flow modality and the incompatibility of XE loss. Then, we analyze and argue that effective motion modeling is essential in WSTAL. Accordingly, we apply GCNs on optical flow to obtain a context-dependent motion prior, termed as \textit{motionness}. Besides, we use it to modulate the video-level classification, yielding a novel Motion-guided Loss. Experiments conducted on three benchmarks including THUMOS'14, ActivityNet v1.2, and ActivityNet v1.3 datasets have validated the state-of-the-art performance of our proposed DMP-Net.

%
\IEEEpeerreviewmaketitle

\ifCLASSOPTIONcaptionsoff
  \newpage
\fi



%

\bibliography{refs.bib}
\bibliographystyle{plain} 

\end{document}